\documentclass[10pt,twocolumn,letterpaper]{article}
\usepackage{cvpr}
\usepackage{epsfig}
\usepackage{graphicx}
\usepackage{amsmath}
\usepackage{amssymb}
\usepackage{listings}

\usepackage{makecell}
\usepackage{multirow}

\usepackage[latin9]{inputenc}
\usepackage[english]{babel}
\usepackage{caption}% <-- added
\usepackage{tabulary}
\usepackage[para]{threeparttable}
\usepackage{array,booktabs,longtable,tabularx}
\usepackage{ltablex}% <-- added
\usepackage{siunitx}% <-- added
\usepackage{caption}% <-- added
\usepackage[flushleft]{threeparttablex}
\usepackage{footnote}

\usepackage{tablefootnote}
\usepackage{url}
\graphicspath{ {./images/} }
\makesavenoteenv{tabular}
\makesavenoteenv{table}

\usepackage[pagebackref=true,breaklinks=true,letterpaper=true,colorlinks,bookmarks=false]{hyperref}
\PassOptionsToPackage{hyphens}{url}

\cvprfinalcopy % *** Uncomment this line for the final submission

\usepackage[numbers,square,comma,sort&compress]{natbib}
\usepackage{textcomp}

\begin{document}

%%%%%%%%% TITLE
\title{Neural Network Compression Framework for fast model inference}

\author{Alexander Kozlov\\
Intel\\
{\tt\small alexander.kozlov@intel.com}
% For a paper whose authors are all at the same institution,
% omit the following lines up until the closing ``}''.
% Additional authors and addresses can be added with ``\and'',
% just like the second author.
% To save space, use either the email address or home page, not both
\and
Ivan Lazarevich\\
Intel\\
{\tt\small ivan.lazarevich@intel.com}
\and
Vasily Shamporov\\
Intel\\
{\tt\small vasily.shamporov@intel.com}
\and
Nikolay Lyalyushkin\\Intel\\
{\tt\small nikolay.lyalyushkin@intel.com}
\and
Yury Gorbachev\\
Intel\\
{\tt\small yury.gorbachev@intel.com}
}

\maketitle
%\thispagestyle{empty}

%%%%%%%%% ABSTRACT
\begin{abstract}
We present a new PyTorch-based framework for neural network compression with fine-tuning named Neural Network Compression Framework (NNCF)\footnote{Available at: \href{https://github.com/openvinotoolkit/nncf}{https://github.com/openvinotoolkit/nncf}}. It leverages recent advances of various network compression methods and implements some of them, namely quantization, sparsity, filter pruning and binarization. These methods allow producing more hardware-friendly models that can be efficiently run on general-purpose hardware computation units (CPU, GPU) or specialized deep learning accelerators. We show that the implemented methods and their combinations can be successfully applied to a wide range of architectures and tasks to accelerate inference while preserving the original model's accuracy. The framework can be used in conjunction with the supplied training samples or as a standalone package that can be seamlessly integrated into the existing training code with minimal adaptations.

\end{abstract}

%%%%%%%%% BODY TEXT
\section{Introduction}
Deep neural networks (DNNs) have contributed to the most important breakthroughs in machine learning over the last ten years \cite{AlexNet,VGG,GNMT,WAVENet}. State-of-the-art accuracy in the majority of machine learning tasks was improved by introducing DNNs with millions of parameters trained in an end-to-end fashion. However, the usage of such models dramatically affected the performance of algorithms because of the billions of operations required to make accurate predictions. Model analysis \cite{zeiler2014,rodriguez2016,CReLU} has shown that a majority of the DNNs have a high level of redundancy, basically caused by the fact that most networks were designed to achieve the highest possible accuracy for a given task without inference runtime considerations. Model deployment, on the other hand, has a performance/accuracy trade-off as a guiding principle. This observation motivated the development of methods to train more computationally efficient deep learning (DL) models, so that these models can be used in real-world applications with constrained resources, such as inference on edge devices. 

Most of these methods can be roughly divided into two categories. The first category contains the neural architecture search (NAS) algorithms \cite{NASNet,MNASNet,PNASNet}, which allow constructing efficient neural networks for a particular dataset and specific hardware used for model inference. The second category of methods aims to improve the performance of existing and usually hand-crafted DL models without much impact on their architecture design. Moreover, as we show in this paper, these methods can be successfully applied to the models obtained using NAS algorithms. One example of such methods is quantization \cite{QuantizationGoogle,PACT}, which is used to transform the model from floating-point to fixed-point representation and allows using the hardware supporting fixed-point arithmetic in an efficient way. The extreme case of quantized networks are binary networks \cite{BNN,XNOR,DOREFA} where the weights and/or activations are represented by one of two available values so that the original convolution/matrix multiplication can be equivalently replaced by XNOR and POPCOUNT operations, leading to a dramatic decrease in inference time on suitable hardware. Another method belonging to this group is based on introducing sparsity into the model weights \cite{SparseCNN,FasterCNNSparsity,SparsityL0} which can be further exploited to reduce the data transfer rate at inference time, or bring a performance speed-up via the sparse arithmetic given that it is supported by the hardware.

In general, any method from the second group can be applied either during or after the training, which adds a further distinction of these methods into post-training methods and methods which are applied in conjunction with fine-tuning. Our framework contains methods that use fine-tuning of the compressed model to minimize accuracy loss.

Shortly, our contribution is a new NNCF framework which has the following important features:
\begin{itemize}
    \item Support of quantization, binarization, sparsity and filter pruning algorithms with fine-tuning.
    \item Automatic model graph transformation in PyTorch -- the model is wrapped and additional layers are inserted in the model graph.
    \item Ability to stack compression methods and apply several of them at the same time.
    \item Training samples for image classification, object detection and semantic segmentation tasks as well as configuration files to compress a range of models.
    \item Ability to integrate compression-aware training into third-party repositories with minimal modifications of the existing training pipelines, which allows integrating NNCF into large-scale model/pipeline aggregation repositories such as {\it MMDetection} \cite{MMDet} or {\it Transformers} \cite{Transformers}.
    \item Hardware-accelerated layers for fast model fine-tuning and multi-GPU training support.
    \item Compatibility with OpenVINO\textsuperscript{TM} Toolkit \cite{OpenVINO} for model inference.
\end{itemize}

It worth noting that we make an accent on production in our work in order to provide a simple but powerful solution for the inference acceleration of neural networks for various problem domains. Moreover, we should also note that we are not the authors of all the compression methods implemented in the framework. For example, binarization and RB sparsity algorithms described below were taken from third party projects by agreement with their authors and integrated into the framework.

\section{Related Work}\label{RelatedWorks}
Currently, there are multiple efforts to bring compression algorithms not only into the research community but towards a wider range of users who are interested in real-world DL applications. Almost all DL frameworks, in one way or another, provide support of compression features. For example, quantizing a model into INT8 precision is now becoming a mainstream approach to accelerate inference with minimum effort. 

One of the influential works here is \cite{QuantizationGoogle}, which introduced the so-called Quantization-Aware Training (QAT) for TensorFlow. This work highlights problems of algorithmic aspects of uniform quantization for CNNs with fine-tuning, and also proposes an efficient inference pipeline based on the instructions available for specific hardware. The QAT is based on the \textit{Fake Quantization} operation which, in turn, can be represented by a pair of \textit{Quantize/Dequantize} operations. The important feature of the proposed software solution is the automatic insertion of the Fake Quantization operations, which makes model optimization more straightforward for the user. However, this approach has significant drawbacks -- namely, increased training time and memory consumption. Another concern is that the quantization method of \cite{QuantizationGoogle} is based on the naive min/max approach and potentially may achieve worse results than the more sophisticated quantization range selection strategies. The latter problem is solved by methods proposed in \cite{PACT}, where quantization parameters are learned using gradient descent. In our framework we use a similar quantization method, along with other quantization schemes, while also providing the ability to automatically insert Fake Quantization operations in the model graph.

Another TensorFlow-based Graffitist framework, which also leverages the training of quantization thresholds \cite{TQT}, aims to improve upon the QAT techniques by providing range-precision balancing of the resultant per-tensor quantization parameters via training these jointly with network weights. This scheme is similar to ours but is limited to symmetric quantization, factor-of-2 quantization scales, and only allows for 4/8 bit quantization widths, while our framework imposes no such restrictions to be more flexible to the end users. Furthermore, NNCF does not perform additional network graph transformations during the quantization process, such as batch normalization folding which requires additional computations for each convolutional operation, and therefore is less demanding to memory and computational resources. 

From the PyTorch-based tools available for model compression, the Neural Network Distiller \cite{distiller} is the well-known one. It contains an implementation of algorithms of various compression methods, such as quantization, binarization, filter pruning, and others. However, this solution mostly focuses on research tasks rather than the application of the methods to real use cases. The most critical drawback of Distiller is the lack of a ready-to-use pipeline from the model compression to the inference on the target hardware.

The main feature of existing compression frameworks is usually the ability to quantize the weights and/or activations of the model from 32 bit floating point into lower bit-width representations without sacrificing much of the model's accuracy. However, as it is now commonly known \cite{SongHan_sparsity}, deep neural networks can also typically tolerate high levels of sparsity, that is, a large proportion of weights or neurons in the network can be zeroed out without much harm to model's accuracy. NNCF allows to produce compressed models that are both quantized and sparsified. The sparsity algorithms implemented in NNCF constitute non-structured network sparsification approaches, i.e. methods that result in sparse weight matrices of convolutional and fully-connected layers with zeros randomly distributed inside the weight tensors. Another approach is the so-called structured sparsity, which aims to prune away whole neurons or convolutional filters \cite{ChannelPruning}. NNCF implements a set of filter pruning algorithms for convolutional neural networks. The non-structured sparsity algorithms generally range from relatively straightforward magnitude-based weight pruning schemes \cite{SongHan_sparsity,MagnitudeSparsity} to more complex approaches such as variational and targeted dropout \cite{VariationalDropout,TargetedDropout} and $L_0$ regularization \cite{RBSparsity}.
%Refer to SparseQuantizer which can give a sparse+quantize model.

\section{Framework Architecture}

NNCF is built on top of the popular PyTorch framework. Conceptually, NNCF consists of an integral core part with a set of compression methods which form the NNCF Python package, and of a set of training samples which demonstrate capabilities of the compression methods implemented in the package on several key machine learning tasks. 

To achieve the purposes of simulating compression during training, NNCF wraps the regular, base full-precision PyTorch model object into a transparent \textit{NNCFNetwork} wrapper. Each compression method acts on this wrapper by defining the following basic components:
\begin{itemize}
    \item \textit{Compression Algorithm Builder} -- the entity that specifies the changes that have to be made to the base model in order to simulate the compression specific to the current algorithm.
    \item \textit{Compression Algorithm Controller} -- the entity that provides access to the compression algorithm parameters and statistics during training (such as the exact quantization bit width of a certain layer in the model, or the level of sparsity in a certain layer).
    \item \textit{Compression Loss}, representing an additional loss function introduced in the compression algorithm to facilitate compression.
    \item \textit{Compression Scheduler}, which can be defined to automatically control the parameters of the compression method during the training process, with updates on a per-batch or per-epoch basis without explicitly using the \textit{Compression Algorithm Controller}.
\end{itemize}

We assume that potentially any compression method can be implemented using these abstractions. For example, the Regularization-Based (RB) sparsity method implemented in NNCF introduces importance scores for convolutional and fully-connected layer weights which are additional trainable parameters. A weight binary mask based on an importance score threshold is added by a specialization of a \textit{Compression Algorithm Builder} object acting on the \textit{NNCFNetwork} object, modifying it in such a manner that during the forward pass the weights of an operation are multiplied by the mask before executing the operation itself. In order to effectively train these additional parameters, RB sparsity method defines an $L_0$-regularization loss which should be minimized jointly with the main task loss, and also specifies a scheduler to gradually increase the sparsity rate after each training epoch.

As mentioned before, one of the important features of the framework is automatic model transformation, i.e. the insertion of the auxiliary layers and operations required for a particular compression algorithm. This requires access to the PyTorch model graph, which is actually not made available by the PyTorch framework. To overcome this problem we patch PyTorch module operations and wrap the basic operators such as \texttt{torch.nn.functional.conv2d} in order to be able to trace their calls during model execution and execute compression-enabling code before and/or after the operator calls.

Another important novelty of NNCF is the support of algorithm stacking where the users can build custom compression pipelines by combining several compression methods. An example of that are the models which are trained to be sparse and quantized at the same time to efficiently utilize sparse fixed-point arithmetic of the target hardware. The stacking/mixing feature implemented inside the framework does not require any adaptations from the user's side. To enable it one only needs to specify the set of compression methods to be applied in the configuration file.

\begin{figure*}[!ht]
\includegraphics[width=1.0\textwidth]{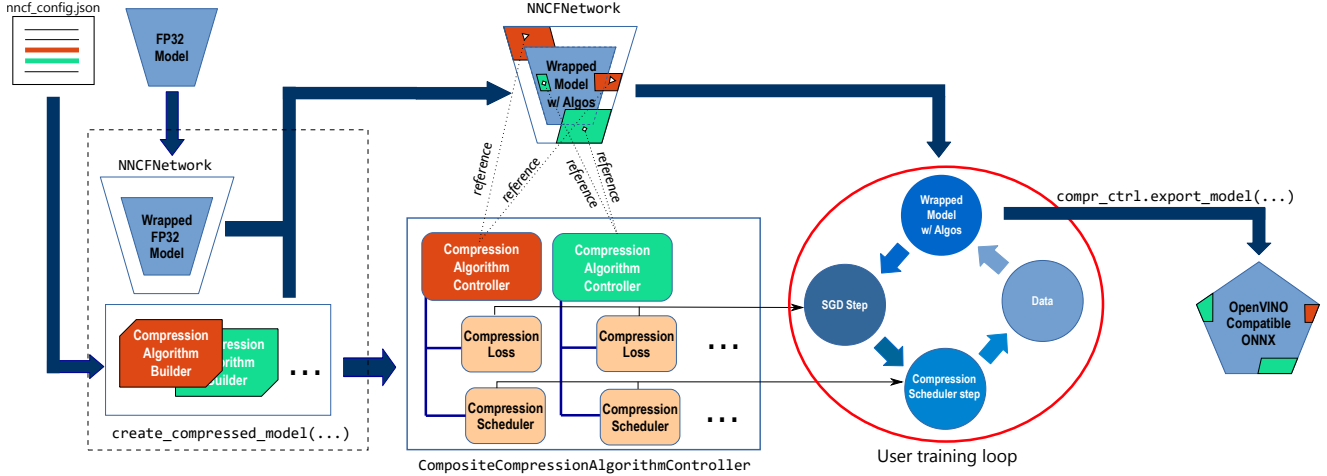}
\caption{A common model compression pipeline with NNCF. The original full-precision model is wrapped and modified to inject compression algorithm functionality into the trained model while leaving the external interface unchanged; this makes it possible to fine-tune the modified model for a pre-defined number of epochs as if it were a regular PyTorch model. Compression algorithm functionality is handled by a Command and control over the specific compression algorithm parameters during training is made available through \texttt{CompressionAlgorithmController} objects.}
\label{fig:nncf_pipeline}
\end{figure*}

Fig.~\ref{fig:nncf_pipeline} shows the common training pipeline for model compression. During the initial step the model is wrapped by the transparent \textit{NNCFNetwork} wrapper, which keeps the original functionality of the model object unchanged so that it can be further used in the training pipeline as if it had not been modified at all. Next, one or more particular compression algorithm builders are instantiated and applied to the wrapped model. The application step produces one or more compression algorithm controllers (one for each compression algorithm) and also the final wrapped model object with necessary compression-related adjustments in place. The wrapped model can then be fine-tuned on the target dataset using either an original training pipeline, or a slightly modified pipeline in case the user decided to apply an algorithm that specifies an additional \textit{Compression Loss} to be minimized or to use a \textit{Compression Scheduler} for automatic compression parameter adjustment during training. The slight modifications comprise, respectively, of a call to the \textit{Compression Loss} object to compute the value to be added to the main task loss (e.g. a cross-entropy loss in case of classification task), and calls to \textit{Compression Scheduler} at regular times (for instance, once per training epoch) to signal that another step in adjusting the compression algorithm parameters should be taken (such as increasing the sparsity rate). As we show in Appendix~\ref{appl:code_modification} any existing training pipeline written with PyTorch can be easily adapted to support model compression using NNCF. After the compressed model is trained we can export it to ONNX format for further usage in the OpenVINO\textsuperscript{TM} \cite{OpenVINO} inference toolkit. 

\section{Compression Methods Overview}
In this section we give an overview of the compression methods implemented in the NNCF framework.
\section*{Quantization}
%The most common numerical format used for storing and computing Deep Learning models has been 32-bit floating point. But it does not meet demand for launching novel computationally cost  models at the edge. It has driven research into developing quantization schemes that converts floating point number to the integer one with lower bit-width. 
The first and most common DNN compression method is quantization. Our quantization approach combines the ideas of QAT~\cite{QuantizationGoogle} and PACT~\cite{PACT} and very close to TQT~\cite{TQT}: we train quantization parameters jointly with network weights using the so-called "fake" quantization operations inside the model graph. But in contrast to TQT, NNCF supports symmetric and asymmetric schemes for activations and weights as well as the support of per-channel quantization of weights which helps quantize even lightweight models produced by NAS, such as EfficientNet-B0.

For all supported schemes quantization is represented by the affine mapping of integers $q$ to real numbers $r$:

\begin{equation}
q=\frac{r}{s} + z \label{eq:quant_design}
\end{equation}

where $s$ and $z$ are quantization parameters. The constant $s$ ("scale factor") is a positive real number, $z$ (zero-point) has the same type as quantized value $q$ and maps to the real value $r=0$. Zero-point is used for asymmetric quantization and provides proper handling of zero paddings. For symmetric quantization it is equal to 0.

\textit{Symmetric quantization.} During the training we optimize the $scale$ parameter that represents range $[{r_{min}}, {r_{max}}]$ of the original signal:

$$
[{r_{min}}, {r_{max}}]=[-scale*\frac{q_{min}}{q_{max}}, scale]
$$

where $[{q_{min}}, {q_{max}}]$ defines quantization range. Zero-point always equal to zero in this case.  Quantization ranges for activation and weights are tailored toward the hardware options available in the OpenVINO\textsuperscript{TM} Toolkit (see Table \ref{table:QRangesForSymmetric}).
Three point-wise operations are sequentially applied to quantize $r$ to $q$: scaling, clamping and rounding.
\begin{equation}
    q = \left\lfloor clamp(\frac{r}{s}; {q_{min}}, {q_{max}})\right \rceil \label{eq:sym_quant}
\end{equation}
\begin{gather*}
    s = \frac{q_{max}}{scale} \\
    clamp(x; a, b) = min(max(x, a), b))
\end{gather*}

where $\left\lfloor\cdot\right \rceil$ denotes the ``bankers'' rounding operation.    

\setlength{\tabcolsep}{4pt}
\begin{table}[]
\centering
\caption{Integer quantization ranges for the symmetric mode at different bit-widths.}
\label{table:QRangesForSymmetric}
\begin{tabular}{l|ccc}
                    & $q_{min}$         & $q_{max}$        \\
\hline
Weights             & $-2^{bits-1}+1$ & $2^{bits-1}-1$ \\
Signed Activation   & $-2^{bits-1}$   & $2^{bits-1}-1$ \\
Unsigned Activation & $0$             & $2^{bits}-1$   \\ 
\hline
\end{tabular}
\end{table}
\textit{Asymmetric quantization.} Unlike symmetric quantization, for asymmetric we optimize boundaries of floating point range (${r_{min}}$, ${r_{max}}$) and use zero-point ($z$) from (\ref{eq:quant_design}).

\begin{equation}
q = \left\lfloor \frac{clamp(r; {r_{min}}, {r_{max}})}{s} + z \right \rceil \label{eq:asym_quant}
\end{equation}
\begin{gather*}
s = \frac{r_{max}-r_{min}}{2^{bits}-1} \\
z = \frac{-r_{min}}{s}
\end{gather*}

In addition we add a constraint to the quantization scheme: floating-point zero should be exactly mapped into an integer within quantization range. This constraint allows efficient implementation of layers with padding. Therefore we tune the ranges before quantization with the following scheme:

\begin{gather*}
l_1 = min(r_{min}, 0) \\
h_1 = max(r_{max}, 0) \\
{z}'= \left\lfloor \frac{-l_1*(2^{bits}-1)}{h_1-l_1} \right \rceil \\
t=\frac{{z}'-2^{bits}+1}{{z}'} \\
h_2=t*l_1 \\
l_2=\frac{h_1}{t} \\
[r_{min},r_{max}] = 
\begin{cases} 
[l_1,h_1], & {z}' \in \{0,2^{bits}-1\} \\ 
[l_1,h_2], & h_2 - l_1 > h_1 - l_2 \\ 
[l_2,h_1], & h_2 - l_1 <= h_1 - l_2
\end{cases}
\end{gather*}

\textit{Comparing quantization modes.} The main advantage of symmetric quantization is simplicity. It does not have a zero-point, which introduces additional logic in hardware. The asymmetric mode, on the other hand, allows fully utilizing quantization ranges, which may potentially lead to better accuracy, especially for quantization lower than 8-bit, as we show in the Table~\ref{table:schemes_comparision}.
\setlength{\tabcolsep}{4pt}
\begin{table}[!ht]
\centering
\caption{Quantization scheme comparison at different bit-widths for weights and activations using MobileNet-V2 on CIFAR-100 dataset. Top-1 accuracy is used as a metric. 32-bit floating point model has an accuracy of 65.53\%.} \label{table:schemes_comparision}
\begin{tabular}{l|ccc}
Scheme  & W8/A8 & W4/A8 & W4/A4\\
\hline
Symmetric & 65.93 & 64.42 & 62.9 \\
Asymmetric & 66.1 & 65.87 & 64.7\\
\hline
\end{tabular}
\end{table}

\textit{Training and inference.} As it was mentioned, quantization is simulated on the forward pass during training by means of \textit{FakeQuantization} operations which perform quantization according to (\ref{eq:sym_quant}) or (\ref{eq:asym_quant}) and dequantization (\ref{eq:dequant_design}) at the same time:
\begin{equation}
    r = s * (q - z) \label{eq:dequant_design}
\end{equation}
FakeQuantization layers are automatically inserted in the model graph. Weights get "fake" quantized before the corresponding operations. Activations are quantized when the preceding layer changes the data type of the tensor, except for basic fusion patterns which correspond to one operation at inference time, such as Conv + ReLU or Conv + BatchNorm + ReLU.

Unlike QAT~\cite{QuantizationGoogle} and TQT~\cite{TQT}, we do not do BatchNorm folding in order to avoid the double computation of convolutions and additional memory consumption which significantly slows down the training. However, to avoid misalignment between BatchNorm statistics during the training and inference we need to use a large batch size (256 samples or more).

%In comparison to QAT, we save training time and reduce memory consumption by not simulating batch normalization during the training and fusing it to convolution biases and quantization parameters of convolution weights. 
%NNCF provides several parameters to customize quantization:
%\begin{itemize}
%    \item \textit{bits} - bits width for representation of quantized values
%    \item \textit{mode} - quantization mode: asymmetric or symmetric
%    \item \textit{signed} - defines quantization range for activations
%    \item \textit{ignored\_scopes} and \textit{target\_scopes} - a list of layers that should be excluded/included from/to the compression
%    \item \textit{quantizable\_subgraph\_patterns} - combinations of layers between which simulation of quantization is not needed
%    \item \textit{num\_init\_steps} - a number of iterations to infer model on part of training dataset to collect minimum and maximum values of weights and  activations. Based on these values quantization ranges are initialized , which promotes more rapid convergence of training.
%\end{itemize}

\begin{table*}[!ht]
\centering
\caption{INT8 quantization results measured in the training framework (PyTorch) -- accuracy metrics on the validation set for original and compressed models.}
\label{table:quantization_results}
\begin{tabular*}{\linewidth}{S @{\extracolsep{\fill}}
                        *{5}{S[group-separator = {,},
                               group-minimum-digits = 6
                                ]}}
\toprule
{Model} & {Dataset} & {Metric type} & {FP32} & {Compressed}  \\
\hline 
{ResNet-50} & {ImageNet} & {top-1 acc.} & {76.13} & {76.03}\\
{Inception-v3} & {ImageNet} & {top-1 acc.} & {77.32} & {78.36}\\
{MobileNet-v1} & {ImageNet} & {top-1 acc.} & {69.6} & {69.75}\\
{MobileNet-v2} & {ImageNet} & {top-1 acc.} & {71.8} & {71.8}\\
{MobileNet-v3 Small} & {ImageNet} & {top-1 acc.} & {67.1} & {66.77}\\
{SqueezeNet v1.1} & {ImageNet} & {top-1 acc.} & {58.19} & {58.16}\\
{SSD300-BN} & {VOC07+12} & {mAP} & {78.28} & {78.18}\\
{SSD512-BN} & {VOC07+12} & {mAP} & {80.26} & {80.32}\\
{UNet} & {Camvid} & {mIoU} & {72.5} & {73.0}\\
{UNet} & {Mapillary Vistas} & {mIoU} & {56.23} & {56.16}\\
{ICNet} & {Camvid} & {mIoU} & {67.89} & {67.78}\\
{BERT-base-chinese} & {XNLI (test, Chinese)} & {top-1 acc.} & {77.68} & {77.02} \\
{BERT-large-uncased-wwm$^*$} & {SQuAD v1.1 (dev)} & {F1} & {93.21} & {92.48}\\
{DistilBERT-base} & {SST-2} & {top-1 acc.} & {91.1} & {90.3} \\
{MobileBERT} & {SQuAD v1.1 (dev)} & {F1} & {89.98} & {89.4}\\
{GPT-2} & {WikiText-2 (raw)} & {Perplexity} & {19.73} & {20.9}\\
\bottomrule
\end{tabular*}
{\raggedright \textsuperscript{*} Whole word masking. \par}
\end{table*}

\textit{Mixed precision quantization}.
Quantization to lower precisions (e.g. 6, 4, 2 bits) is an efficient way to accelerate the inference of neural networks and achieve significant weight compression. Although NNCF supports quantization with an arbitrary number of bits representing weight and activation values, choosing ultra-low bit-widths could noticeably affect the model's accuracy. A good trade-off between accuracy and performance is achieved by assigning different precisions to different layers. NNCF utilizes the HAWQ-v2~\cite{HAWQv2} method to automatically choose the optimal mixed-precision configuration by taking into account the sensitivity of each layer, i.e. how much the low bit-width quantization of each layer decreases the accuracy of the model. The most sensitive layers are kept at higher precision. The sensitivity of the $i$-th layer is calculated by multiplying the average Hessian trace with the L2 norm of weight quantization perturbation

\begin{equation}
\overline{Tr}(H_{i})*\left\|Q(W_{i})-W_{i}\right\|^2_2
\end{equation}

where the trace is estimated using the randomized Hutchinson algorithm~\cite{Hutchinson}.
The sum of the sensitivities for each layer forms a metric which serves as a proxy to the accuracy of the compressed model: the lower the metric, the more accurate should the corresponding mixed precision model be on the validation dataset. To find the optimal trade-off between accuracy and performance of the mixed precision model we also compute a compression ratio -- the ratio between bit complexity of a fully INT8 model and mixed-precision lower bit-width one. The bit complexity of the model is a sum of bit complexities for each quantized layer, which are defined as a product of the layer FLOPs and the quantization bit-width. The optimal configuration is found by calculating the sensitivity metric and the compression ratio for all possible bit-width settings and selecting the one with the minimal metric value among all configurations with a compression ratio below the specified threshold.
To avoid the exponential search procedure, we apply the following restriction: layers with a small average Hessian trace value are quantized to lower bit-widths and vice versa. 

\begin{table*}[!ht]
\centering
\caption{Mixed precision quantization results measured in the training framework (PyTorch) -- accuracy metrics on validation set for original and compressed models.}
\label{table:mixed_precision_results}
\begin{tabular*}{\linewidth}{S @{\extracolsep{\fill}}
                        *{5}{S[group-separator = {,},
                               group-minimum-digits = 6
                                ]}}
\toprule
{Model} & {Dataset} & {Metric type} & {FP32} & {Compressed}  \\
\hline 
{ResNet-50} \\ {44.8\% INT8 / 55.2\% INT4} & {ImageNet} & {top-1 acc.} & {76.13} & {76.3}\\
\hline 
{MobileNet-v2} \\ {46.6\% INT8 / 53.4\% INT4} & {ImageNet} & {top-1 acc.} & {71.8} & {70.89}\\
\hline 
{SqueezeNet-v1.1} \\ {54.7\% INT8 / 45.3\% INT4} & {ImageNet} & {top-1 acc.} & {58.19} & {58.85}\\
\bottomrule
\end{tabular*}
\end{table*}

\section*{Binarization}\label{Binarization}

Currently, NNCF supports binarizing weights and activations of 2D convolutional PyTorch layers (Conv2D layers).

Weight binarization can be done either via XNOR binarization \cite{rastegari2016xnor} or via DoReFa binarization \cite{zhou2016dorefa} schemes. For DoReFa binarization the scale of binarized weights for each convolution operation is calculated as the mean of absolute values of non-binarized convolutional filter weights, while for XNOR binarization each convolutional operation will have scales that are calculated in the same manner, but {\it per-input channel} for the convolutional filter. 

Activation binarization is implemented via binarizing inputs of the convolutional layers in the following way:

\begin{equation}
\label{eq:binarize_act}
out = s * H(in - s*t)
\end{equation}

where $in$ are the non-binarized activation values, $out$ - binarized activation values, $H(x)$ is the Heaviside step function and $s$ and $t$ are trainable parameters corresponding to binarization scale and threshold respectively. The thresholds $t$ are trained separately for each output activation channel dimension.

It is usually not recommended to binarize certain layers of CNNs - for instance, the input convolutional layer, the fully connected layer and the convolutional layer directly preceding it or the ResNet ``downsample'' layers. NNCF allows picking an exact subset of layers to be binarized via the layer allowlist/denylist mechanism as in other NNCF compression methods. 

Finally, training binarized networks requires a special scheduling of the training process, tailored specifically for each model architecture. NNCF samples demonstrate binarization of a ResNet-18 architecture pre-trained on ImageNet using a four-stage process, where each stage taking a certain number of fine-tuning epochs:
\begin{itemize}
    \item Stage 1: the network is trained without any binarization,
    \item Stage 2: the training continues with binarization enabled for activations only,
    \item Stage 3: binarization is enabled both for activations and weights,
    \item Stage 4: the optimizer learning rate, which had been kept constant at previous stages, is decreased according to a polynomial law, while weight decay parameter of the optimizer is set to 0.
\end{itemize}
 The configuration files for the NNCF binarization algorithm allow controlling the stage durations of this training schedule. The above training pipeline allows getting state-of-the-art accuracy for a ResNet-18 model on ImageNet (see Table \ref{table:binarization_results}).
 
\begin{table*}[!ht]
\centering
\caption{Binarization results measured in the training framework (PyTorch) -- accuracy metrics on validation set for original and compressed models. The models were trained and validated on the ImageNet dataset.}
\label{table:binarization_results}
\begin{tabular*}{\linewidth}{S @{\extracolsep{\fill}}
                        *{5}{S[group-separator = {,},
                               group-minimum-digits = 6
                                ]}}
\toprule
{Model} & {Weight / activation bin type} & {\% ops binarized} & {FP32} & {Compressed} \\
\hline 
{ResNet-18} & {XNOR / scale-threshold} & {92.4} & {69.75} & {61.71}\\
{ResNet-18} & {DoReFa / scale-threshold} & {92.4} & {69.75} & {61.58}\\
\bottomrule
\end{tabular*}
\end{table*}

\section*{Sparsity}

NNCF supports two non-structured weight pruning algorithms: i.) a simple magnitude-based sparsity training scheme, and ii.) $L_0$ regularization-based training, which is a modification of the method proposed in \cite{RBSparsity}. It has been argued \cite{SparsityBenchmark} that complex approaches to sparsification like $L_0$ regularization produce inconsistent results when applied to large benchmark dataset (e.g. ImageNet for classification, as opposed to e.g. CIFAR-100) and that magnitude-based sparsity algorithms provide comparable or better results in these cases. However, we found out in our experiments that the regularization-based (RB) approach to sparsity outperforms the simple magnitude-based method for several classification models trained on ImageNet, achieving higher accuracy for the same sparsity level value for several models (e.g. for MobileNet-V2). Hence, both methods could be used in different contexts, with RB sparsity requiring tuning of the training schedule and a longer training procedure, but ultimately producing better results for certain tasks. We briefly describe the details of both network sparsification algorithms implemented in NNCF below.

\textit{Magnitude-based sparsity.} In the magnitude-based weight pruning algorithm, the normalized magnitude of each weight is used as a measure of its importance (normalization is done on a per-layer basis). In the NNCF implementation of magnitude-based sparsity, a certain schedule for the desired sparsity level $SL$ over the training process is defined, and a threshold value is calculated each time $SL$ is changed by the compression scheduler. Weights that are lower than the calculated threshold value are then zeroed out. The compression scheduler can be set to increase the sparsity rate from an initial to a final level over a certain number of training epochs. The dynamics of sparsity level increase during training are adjustable and support $polynomial$, $exponential$, $adaptive$ and $multistep$ modes.

%The compression scheduler in case of sparsity algorithms has the following adjustable parameters which define the training procedure: 
%\begin{itemize}
%    \item $sparsity\_training\_steps$ -- an overall number of epochs that are used to train sparsity (in general case it is different from the total number of training epochs). After this epoch, the sparsity masks will be frozen, so only the model parameters (weights) will be fine-tuned
%    \item $schedule$ -- a type of scheduler that is used to increase the sparsity rate from $sparsity\_init$ to $sparsity\_target$. Can be one of these values: i) $polynomial$, ii) $exponential$, iii) $adaptive$ or iv) $multistep$. The $multistep$ option is used to define the step-wise dynamics of target sparsity rate as a function of training iterations with a list of $steps$ (epoch numbers at which target sparsity rate is changed) and $sparsity_levels$ for corresponding training intervals.
%    \item $ignored\_scopes$ -- a list of layers that should be excluded from the compression (optional).
%\end{itemize}

\textit{Regularization-based sparsity.} In the RB sparsity algorithm, a \textit{complexity} loss term is added to the total loss function during training, defined as
\begin{equation}
    L_{reg} = \left( \sum_{j=1}^{|\theta|} \frac{\mathbb{I}[\theta_j \neq 0]}{|\theta|} - (1 - SL) \right)^2
\end{equation}
where $|\theta|$ is the number of network parameters and $SL$ is the desired sparsity level (percentage of zero weights) of the network. Note that the above regularization loss term penalizes networks with sparsity levels both lower and higher than the defined $SL$ level. Following the derivations in \cite{RBSparsity}, in order to make the $L_0$ loss term differentiable, the model weights are reparametrized as follows: 
\begin{gather*}
    \theta_j = \hat{\theta_j} z_j, \\
    \text{where } z_j \sim [ \text{sigmoid}(s_j + \text{logit}(u)) > 0.5 ] \\
    \text{and } u \sim U(0, 1)
\end{gather*}
where $z_j$ is the stochastic binary gate, $z_j \in \{0, 1\}$ . It can be shown that the above formulation is equivalent to $z_j$ being sampled from the Bernoulli distribution $B(p_j)$ with probability parameter $p_j = \text{sigmoid}(s_j)$. Hence, ${s_j}$ are the trainable parameters which control whether the weight is going to be zeroed out at test time (which is done for $p_j > 0.5$). 

On each training iteration, the set of binary gate values ${z_j}$ is sampled once from the above distribution and multiplied with network weights. In the Monte Carlo approximation of the loss function in \cite{RBSparsity}, the mask of binary gates is generally sampled and applied several times per training iteration, but single mask sampling is sufficient in practice (as shown in \cite{RBSparsity}). The expected $L_0$ loss term was shown to be proportional to the sum of probabilities of gates $z_j$ being non-zero \cite{RBSparsity}, which in our case results in the following expression
\begin{equation}
    L_{reg} = \left( \sum_{j=1}^{|\theta|} \frac{\text{sigmoid}(s_j)}{|\theta|} - (1 - SL) \right)^2
\end{equation}
To make the error loss term (e.g. cross-entropy for classification) differentiable w.r.t $s_j$, we treat the threshold function $t(x) = [x > c]$ as a straight-through estimator (i.e. $dt/dx = 1$).

\begin{table*}[!ht]
\centering
\caption{Sparsification+quantization results measured in the training framework (PyTorch) -- accuracy metrics on the validation set for the original and compressed models. "RB" stands for the regularization-based sparsity method and "Mag." denotes the simple magnitude-based pruning method.}
\label{table:sparsity_results}
\begin{tabular*}{\linewidth}{S @{\extracolsep{\fill}}
                        *{5}{S[group-separator = {,},
                              group-minimum-digits = 6
                                ]}}
\toprule
{Model} & {Dataset} & {Metric type} & {FP32} & {Compressed} \\
\hline 
{ResNet-50} \\ {INT8 w/ 60\% of sparsity (RB)} & {ImageNet} & {top-1 acc.} & {76.13} & {75.2}\\
\hline
{Inception-v3} \\ {INT8 w/ 60\% of sparsity (RB)} & {ImageNet} & {top-1 acc.} & {77.32} & {76.8}\\
\hline
{MobileNet-v2} \\ {INT8 w/ 51\% of sparsity (RB)} & {ImageNet} & {top-1 acc.} & {71.8} & {70.9}\\
\hline
{MobileNet-v2} \\ {INT8 w/ 70\% of sparsity (RB)}& {ImageNet} & {top-1 acc.} & {71.8} & {70.1}\\
\hline
{SSD300-BN} \\ {INT8 w/ 70\% of sparsity (Mag.)} & {VOC07+12} & {mAP} & {78.28} & {77.94}\\
\hline
{SSD512-BN} \\ {INT8 w/ 70\% of sparsity (Mag.)} & {VOC07+12} & {mAP} & {80.26} & {80.11}\\
\hline
{UNet} \\ {INT8 w/ 60\% of sparsity (Mag.)} & {CamVid} & {mIoU} & {72.5} & {73.27}\\
\hline
{UNet} \\ {INT8 w/ 60\% of sparsity (Mag.)} & {Mapillary} & {mIoU} & {56.23} & {54.30}\\
\hline
{ICNet} \\ {INT8 w/ 60\% of sparsity (Mag.)} & {CamVid} & {mIoU} & {67.89} & {67.53}\\
\bottomrule
\end{tabular*}
\end{table*}

\section*{Filter pruning}

NNCF also supports structured pruning for convolutional neural networks in the form of filter pruning. The filter pruning algorithm zeroes out the output filters in convolutional layers based on a certain filter importance criterion \cite{wen2016learning}. NNCF implements three different criteria for filter importance: i.) L1-norm, ii.) L2-norm and iii.) geometric median. The geometric median criterion is based on the finding that if a certain filter is close to the geometric median of all the filters in the convolution, it can be well approximated by a linear combination of other filters, hence it should be removed. To avoid the expensive computation of the geometric median value for all the filters, we use the following approximation for the filter importance metric \cite{MedianPruning}:
\begin{equation}
    G(F_i) = \sum_{i \in \{1,...,n\}, i \neq j} || F_i - F_j ||_2
\end{equation}
where $F_i$ is the $i$-th filter in a convolutional layer with $n$ output filters. That is, filters with the lowest average L2 distance to all the other filters in the convolutional layer are regarded as less important and eventually pruned away. 
The same proportion of filters is pruned for each layer based on the global pruning rate value set by the user. We found that the geometric median criterion gives a slightly better accuracy for the same fine-tuning pipeline compared to the magnitude-based pruning approach (see Table \ref{table:filter_pruning_results}). We were able to produce a set of ResNet models with 30\% of filters pruned in each layer and a top-1 accuracy drop lower than 1\% on the ImageNet dataset. 

The fine-tuning pipeline for the pruned model is determined by the user-configurable pruning scheduler. First, the original model is fine-tuned for a specified amount of epochs, then a certain target percentage of filters with the lowest importance scores is pruned (zeroed out). At that point in the training pipeline, the pruned filters might be frozen for the rest of the model fine-tuning procedure (this approach is implemented in the \textit{baseline} pruning scheduler). An alternative tuning pipeline is implemented by the \textit{exponential} scheduler, which does not freeze the zeroed out filters until the final target pruning rate is achieved in the model. The initial pruning rate is set to a low value and is increased each epoch according to the exponentially increasing profile. The subset of the filters to be pruned away in each layer is determined every time the pruning rate is changed by the scheduler.

\begin{table*}[!ht]
\centering
\caption{Filter Pruning results measured in the training framework -- accuracy metrics on the validation set for the original and compressed models. The pruning rate for all the reported models was set to 30\%.}
\label{table:filter_pruning_results}
\begin{tabular*}{\linewidth}{S @{\extracolsep{\fill}}
                        *{5}{S[group-separator = {,},
                              group-minimum-digits = 6
                                ]}}
\toprule
{Model} & {Criterion} & {Dataset} & {Metric type} & {FP32} & {Compressed} \\
\hline 
{ResNet-50} & {Magnitude} & {ImageNet} & {top-1 acc.} & {76.13} & {75.7}\\
{ResNet-50} & {Geometric median} & {ImageNet} & {top-1 acc.} & {76.13} & {75.7}\\
{ResNet-34} & {Magnitude} & {ImageNet} & {top-1 acc.} & {73.31} & {75.54}\\
{ResNet-34} & {Geometric median} & {ImageNet} & {top-1 acc.} & {73.31} & {72.62}\\
{ResNet-18} & {Magnitude} & {ImageNet} & {top-1 acc.} & {69.76} & {68.73}\\
{ResNet-18} & {Geometric median} & {ImageNet} & {top-1 acc.} & {69.76} & {68.97}\\
\bottomrule
\end{tabular*}
\end{table*}

Importantly, the pruned filters are removed from the model (not only zeroed out) when it is being exported to the ONNX format, so that the resulting model actually has fewer FLOPS and inference speedup can be achieved. This is done using the following algorithm: first, channel-wise masks from the pruned layers are propagated through the model graph. Each layer has a corresponding function that can compute the output channel-wise masks given layer attributes and input masks (in case mask propagation is possible). After that, the decision on whether to prune a certain layer is made based on whether further operations in the graph can accept the pruned input. Finally, filters are removed according to the computed masks for the operations that can be pruned.

\section{Results}

% Some of the results for different compression methods were disclosed in the corresponding sections.

We show the INT8 quantization results for a range of models and tasks including image classification, object detection, semantic segmentation and several Natural Language Processing (NLP) tasks in Table \ref{table:quantization_results}. The original Transformer-based models for the NLP tasks were taken from the HuggingFace's Transformers repository \cite{Transformers} and NNCF was integrated into the corresponding training pipelines as an external package. Table~\ref{table:efficient_net_results} reports compression results for EfficientNet-B0, which gives best combination of accuracy and performance on the ImageNet dataset. We compare the accuracy values between the original floating point model (76.84\% of top-1) and a compressed one for different compression configurations; top-1 accuracy drop lower than 0.5\% can be observed in most cases.

To go beyond the single-precision INT8 quantization, we trained a range of models quantized to lower bit-widths (see Table \ref{table:mixed_precision_results}). We were able to train several models (ResNet50, MobileNet-v2 and SquezeNet1.1) with approximately half of the model weights (and corresposding input activations) quantized to 4 bits (while the rest of weights and activations were quantized to 8 bits) and a top-1 accuracy drop less than 1\% on the ImageNet dataset. We also trained a binarized ResNet-18 model according to the pipeline described in section \ref{Binarization}. Table \ref{table:binarization_results} presents the results of binarizing ResNet-18 with either XNOR or DoReFa weight binarization and scale-threshold activation binarization (see Eq. \ref{eq:binarize_act}).

We also trained a set of convolutional neural network models with both weight sparsity and quantization algorithms in the compression pipeline (see Table \ref{table:sparsity_results}). To extend the scope of trainable models and to validate that NNCF could be easily combined with existing PyTorch-based training pipelines, we also integrated NNCF with the popular \textit{mmdetection} object detection toolbox \cite{MMDet}. As a result, we were able to train INT8-quantized and INT8-quantized+sparse object detection models available in \textit{mmdetection} on the challenging COCO dataset and achieve a less than 1 mAP point drop for the COCO-based mAP evaluation metric. Specific results for compressed RetinaNet and Mask-RCNN models are shown in Table \ref{table:mmdetection_results}.

\setlength{\tabcolsep}{4pt}
\begin{table}[!ht]
\centering
\caption{Accuracy top-1 results for INT8 quantization of EfficientNet-B0 model on ImageNet measured in the training framework} 
\label{table:efficient_net_results}
\begin{tabular}{l|ccc}
Model  & Accuracy drop\\
\hline
{All per-tensor symmetric} &  {0.75}\\
{All per-tensor asymmetric} & {0.21}\\
{Per-channel weights asymmetric} & {\textbf{0.17}}\\
{All per-tensor asymmetric} \\ {w/ 31\% of sparsity} & {0.35}\\
\hline
\end{tabular}
\end{table}

The compressed models were further exported to ONNX format suitable for inference with the OpenVINO\textsuperscript{TM} toolkit. The performance results for the original and compressed models as measured in OpenVINO\textsuperscript{TM} are shown in Table ~\ref{table:performace_results}.

\setlength{\tabcolsep}{4pt}
\begin{table}[!ht]
\centering
\caption{Relative performance/accuracy results with OpenVINO\textsuperscript{TM} 2020.1 on Intel$^{\tiny{\textregistered}}$ Xeon$^{\tiny{\textregistered}}$ Gold 6230 Processor.} \label{table:performace_results}
\begin{tabular}{l|cc}
Model  & Accuracy drop (\%) & Speed up\\
\hline
MobileNet v2 INT8 & 0.44 & 1.82x\\
ResNet-50 v1 INT8 & -0.34 & 3.05x\\
Inception v3 INT8 & -0.62 & 3.11x\\
SSD-300 INT8 & -0.12 & 3.31x\\
UNet INT8 & -0.5 & 3.14x\\
ResNet-18 XNOR & 7.25 & 2.56x \\
\hline
\end{tabular}
\end{table}

%MobileNet v2 INT8 & 71.81 & 71.33 & -0.44 & 1.82\\
%ResNet-50 v1 INT8 & 76.13 & 76.47 & 0.34 & 3.05\\
%Inception v3 INT8 & 77.69 & 78.31 & 0.62 & 3.11\\
%SSD-300 INT8 & 78.02 & 78.14 & 0.12 & 3.31\\
%ResNet-18 XNOR & 68.96 & 61.71 & -7.25 & 2.56 \\

\setlength{\tabcolsep}{4pt}
\begin{table}[!ht]
\centering
\caption{Validation set metrics for original and compressed $mmdetection$ models. Shown are bounding box mAP values for models trained and tested on the COCO dataset.} \label{table:mmdetection_results}
\begin{tabular}{l|ccc}
Model & FP32 & Compressed\\
\hline
RetinaNet-ResNet50-FPN INT8 & 35.6 & 35.3\\
\hline
{RetinaNet-ResNeXt101-}\\{64x4d-FPN INT8} & 39.6 & 39.1\\
\hline
{RetinaNet-ResNet50-FPN}\\{INT8+50\% sparsity} & 35.6 & 34.7\\
\hline
Mask-RCNN-ResNet50-FPN INT8 & 37.9 & 37.2\\
\end{tabular}
\end{table}

\section{Conclusions}
In this work we presented the new NNCF framework for model compression with fine-tuning. It supports various compression methods and allows combining them to get more lightweight neural networks. We paid special attention to usability aspects and simplified the compression process setup as well as approbated the framework on a wide range of models and tasks. Models obtained with NNCF show state-of-the-art results in terms of accuracy-performance trade-off. The framework is compatible with the OpenVINO\textsuperscript{TM} inference toolkit which makes it attractive to apply the compression to real-world applications. We are constantly working on developing new features and improvement of the current ones as well as adding support of new models.

{\small
\bibliographystyle{ieeetr}
\bibliography{arxiv_version}
}

\appendix
\section{Appendix} \label{appl:code_modification}

Described below are the steps required to modify an existing PyTorch training pipeline in order for it to be integrated with NNCF. The described use case implies there exists a PyTorch pipeline that reproduces model training in floating point precision and a pre-trained model snapshot. The objective of NNCF is to simulate model compression at inference time in order to allow the trainable parameters to adjust to the compressed inference conditions, and then export the compressed version of the model to a format suitable for compressed inference. Once the NNCF package is installed, the user needs to introduce minor changes to the training code to enable model compression. Below are the steps needed to modify the training pipeline code in PyTorch:
\begin{itemize}
    \item Add the following imports in the beginning of the training sample right after importing PyTorch:
    \begin{lstlisting}[language=Python,basicstyle=\ttfamily\small]

import nncf  # Important - should be  
             # imported directly after 
             # torch
from nncf import create_compressed_model, 
                 NNCFConfig, 
                 register_default_init_args

    \end{lstlisting}
    \item Once a model instance is created and the pre-trained weights are loaded, the model can be compressed using the helper methods. Some compression algorithms (e.g. quantization) require arguments (e.g. the \texttt{train\_loader} for your training dataset) to be supplied to the \texttt{initialize()} method at this stage as well, in order to properly initialize compression modulse parameters related to its compression (e.g. scale values for FakeQuantize layers):
    \begin{lstlisting}[language=Python,basicstyle=\ttfamily\small]
# Instantiate your uncompressed model
from torchvision.models.resnet import 
resnet50
model = resnet50()

# Load a configuration file to specify 
# compression
nncf_config = NNCFConfig.from_json(
"resnet50_int8.json")

# Provide data loaders for compression 
# algorithm initialization, if necessary
nncf_config = register_default_init_args
(nncf_config, train_loader, loss_criterion)

# Apply the specified compression 
# algorithms to the model
comp_ctrl, compressed_model = 
create_compressed_model(model,
nncf_config)

    \end{lstlisting}
    where \texttt{resnet50\_int8.json} in this case is a JSON-formatted file containing all the options and hyperparameters of compression methods (the format of the options is imposed by NNCF). 
    \item At this stage the model can optionally be wrapped with \texttt{DataParallel} or\newline\texttt{DistributedDataParallel} classes for multi-GPU training. In case distributed training is used, call the \texttt{compression\_algo.distributed()} method after wrapping the model with \texttt{DistributedDataParallel} to signal the compression algorithms that special distributed-specific internal handling of compression parameters is required.
    
    \item The model can now be trained as a usual torch.nn.Module to fine-tune compression parameters along with the model weights. To completely utilize NNCF functionality, you may introduce the following changes to the training loop code:
    
    1) after model inference is done on the current training iteration, the compression loss should be added to the main task loss such as cross-entropy loss:
    \begin{lstlisting}[language=Python,basicstyle=\ttfamily\small]
compression_loss = comp_ctrl.loss()
loss = cross_entropy_loss + compression_loss
    \end{lstlisting}
    
    2) the compression algorithm schedulers should be made aware of the batch/epoch steps, so add \texttt{comp\_ctrl.scheduler.step()} calls after each training batch iteration and \texttt{comp\_ctrl.scheduler.epoch\_step()} calls after each training epoch iteration.
    
   \item When done finetuning, export the model to ONNX by calling a compression controller's dedicated method, or to PyTorch's .pth format by using the regular \texttt{torch.save} functionality:
    \begin{lstlisting}[language=Python,basicstyle=\ttfamily\small]
# Export to ONNX or .pth when done fine-tuning
comp_ctrl.export_model("compressed_model.onnx")
torch.save(compressed_model.state_dict(), 
           "compressed_model.pth")

    \end{lstlisting}
\end{itemize}

\end{document}